\documentclass[a4paper,fleqn]{cas-dc}

\usepackage[authoryear]{natbib}
\usepackage{cas-common}
\usepackage{times}
\usepackage{xcolor}
\usepackage{tabularx}

\def\tsc#1{\csdef{#1}{\textsc{\lowercase{#1}}\xspace}}
\tsc{WGM}
\tsc{QE}

\usepackage[final]{changes}

\color{black}

\shorttitle{GIT-Mol: A Multi-modal Large Language Model for Molecular Science with Graph, Image, and Text}

\title [mode = title]{GIT-Mol: A Multi-modal Large Language Model for Molecular Science with Graph, Image, and Text}                      
\author[1,2]{Pengfei Liu}

\affiliation[1]{organization={Peng Cheng Laboratory},
    city={Shenzhen},
    postcode={518055}, 
    state={Guangdong Province},
    country={China}}

\author[1]{Yiming Ren}

\author[2]{Jun Tao}

\author[1]{Zhixiang Ren}
\cormark[1]
\affiliation[2]{organization={School of Computer Science and Engineering, Sun Yat-Sen University},
    city={Guangzhou},
    postcode={510006}, 
    state={Guangdong Province},
    country={China}}
    
\cortext[cor1]{Corresponding author at: Peng Cheng Laboratory, Shenzhen, 518055, China}
\cortext[cor2]{E-mail addresses: renzhx@pcl.ac.cn}

\begin{document}
\begin{abstract}
Large language models have made significant strides in natural language processing, enabling innovative applications in molecular science by processing textual representations of molecules.
However, most existing language models cannot capture the rich information with complex molecular structures or images. 
In this paper, we introduce \textbf{GIT-Mol}, a multi-modal large language model that integrates the \textbf{G}raph, \textbf{I}mage, and \textbf{T}ext information.
To facilitate the integration of multi-modal molecular data, we propose \textbf{GIT-Former}, a novel architecture that is capable of aligning all modalities into a unified latent space.
We achieve a 5\%-10\% accuracy increase in properties prediction and a 20.2\% boost in molecule generation validity compared to the baselines.
\added{With the any-to-language molecular translation strategy, our model has the potential to perform more downstream tasks, such as compound name recognition and chemical reaction prediction.}
\end{abstract}

\begin{keywords}
Molecular Representation \sep
Molecule Generation \sep
Large Language Model \sep
Multi-modality
\end{keywords}

\maketitle

\section{Introduction}

Molecular science covers a broad spectrum of fields that study the structures, properties, and interactions of molecules. It is a interdisciplinary field that draws on chemistry, physics, biology, and computer science.
Molecular science is pivotal in drug discovery applications, such as target identification and validation, structure-based drug design, and side effect prediction.
However, most existing methods of discovering new molecules or tweaking existing ones can be time-consuming, expensive, and prone to failure \citep{rodrigues2016counting}. 
More recently, computational methods have shown significant advantages in molecule generation and tweaking \citep{bilodeau2022generative}. These techniques enable rapid identification and optimization of potential drug candidates.
\added{However, these computational methods are limited by substantial computational demands.}

Fortunately, artificial intelligence (AI) and deep learning have emerged as powerful tools for molecular science. These technologies can potentially revolutionize the field by improving speed, accuracy, and scalability in molecular discovery and understanding. Large Language Models (LLMs) have made significant progress in Natural Language Processing (NLP) and molecular science. MolT5 \citep{edwards2022translation}, based on the T5 \citep{raffel2020exploring}, which includes capabilities of molecule captioning (Mol2Cap) and text-based molecule generation (Cap2Mol). LLMs like MolT5 help describe molecules in words and generate structures from the text. However, these text-to-text models can not fully use the advantages of molecular structure data and understand molecular images.
To fuse and understand multi-modal data, the multi-modal large language models (MLLMs) like CLIP \citep{radford2021learning}, ALIGN \citep{jia2021scaling}, and BEIT-3 \citep{wang2022image} have laid the groundwork for adaptive learning across image-text modalities. In molecular science, SwinOCSR \citep{xu2022swinocsr} is designed for image recognition, which significantly aids in document comprehension and multi-modal drug discovery database construction \citep{wang2022multi}. In addition, MoleculeSTM \citep{liu2022multi} and MoMu \citep{su2022molecular} can combine the Simplified Molecular Input Line Entry System (SMILES) \citep{weininger1988smiles} and graph representations for molecular property prediction tasks.

\begin{figure*}[t!]
\centering
\includegraphics[width=\textwidth]{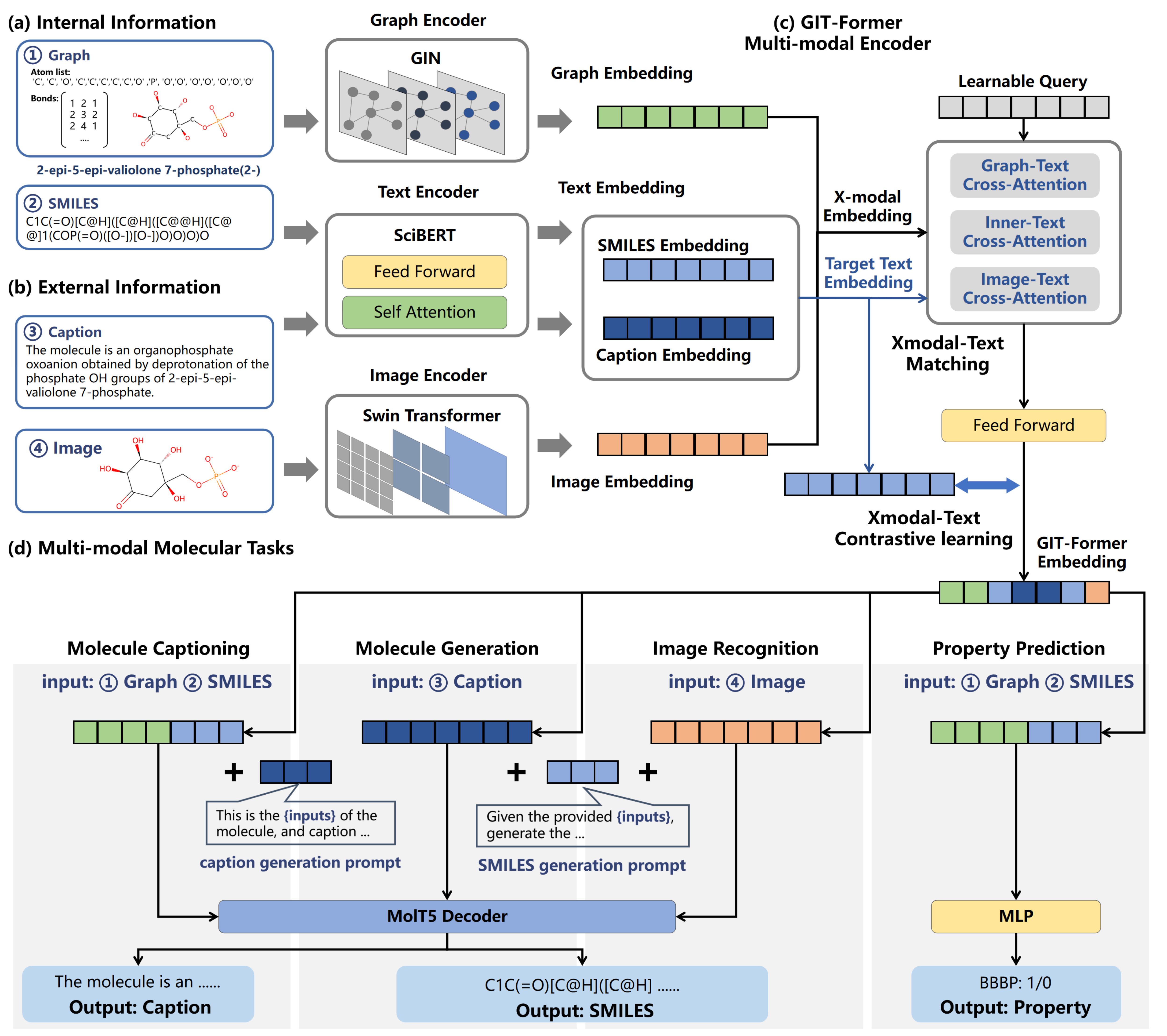}
\caption{\textbf{An overview of GIT-Mol}. \textbf{(a) Internal Information}, including sequence and graph structure representations, emphasizes inherent chemical properties and simple topology; \textbf{(b) External Information}, e.g., images and text descriptions, provide richer details and help the human understanding; \textbf{(c) GIT-Former Multi-modal Encoder}, architecture and Pre-train Strategy of GIT-Former, GIT-Former aligns graph, image, and text with the target text modality (SMILES strings or captions) using self-attention and cross-attention. The learnable queries interact with each other and the various modalities through these attention layers. Xmodal-Text Matching (XTM) and Xmodal-Text Contrastive Learning (XTC) represent our self-supervised learning strategies tailored for specific modalities (X) and target text modalities; \textbf{(d) Multi-modal Molecular Tasks}, in cross-modal tasks, GIT-Former generates different Embeddings based on various inputs, which MolT5 then decodes into the target text modality and the MLP model for property prediction tasks.}
\label{fig:Conceptual-GIT-MOL}
\end{figure*}

\added{
Existing language models excel in processing textual molecular data, but processing molecular graphs and images is difficult, typically relying on processed vector encodings. 
This limitation highlights the need for a modality data alignment process, where multi-modal models outperform language models by integrating diverse data types for enhanced feature representation.
Moreover, the scalability of molecular representation and generation models remains a significant challenge. Models capable of effectively fusing three or more modalities are scarce, and the complexity of integrating these diverse modalities, often with missing or incomplete data, calls for advanced modeling techniques.
}

\added{To address the significant scalability challenges in molecular science and to harness the potential of vast quantities of unlabeled multi-modal data, we have developed GIT-Mol, a robust neural network architecture. 
At the core of GIT-Mol is the GIT-Former, which incorporates an advanced cross-attention mechanism. 
It is capable of mapping data from various modalities into a unified multi-modal representation.
This approach not only enhances the model's ability to process and integrate diverse modalities efficiently but also effectively bridges the scalability gap, providing a powerful tool for various molecular applications.}
Our main contributions are as follows:

\begin{itemize}
\item To utilize the available large amount of unlabeled multi-modal data, we develop a specialized multi-modal large language model, \textbf{GIT-Mol} (700M), to cover all three modalities in molecular science (graph, image, and text) for molecule generation, molecule captioning, molecular image recognition, and molecular property prediction.
\item We present \textbf{GIT-Former}, a novel modality mixer designed with a cross-attention mechanism, enabling seamless fusion of three molecular modalities. Each fusion strategy within GIT-Former operates at the molecule level, ensuring optimal flexibility and scalability. The effectiveness of our multi-modal model is demonstrated in our extensive ablation studies, which provides an improvement of 10\%-15\% over the single-modality models.
\item The quantitative evaluations indicate that GIT-Mol exhibits commendable performance and surpasses state-of-the-art on certain metrics, as it outperforms the baseline 20\% in the cross-modal molecule generation tasks regarding the validity of the generated molecules and by 5\%-10\% in molecular property prediction tasks.
\end{itemize}

\added{
In the subsequent sections, Section \ref{sec:Related Works} primarily introduces multi-modal models and applications in molecular science. Section \ref{sec:Methodology} introduces the principles of our GIT-Mol model. Section \ref{sec:Results} focuses on the experimental setup and results for downstream tasks of the model. Sections \ref{sec:Discussion}, \ref{sec:Limitations}, and \ref{sec:Conclusions} respectively discuss the features of our model, the limitations, and the conclusions.
}

\section{Related Works}
\label{sec:Related Works}

\subsection{Multi-modal Large Language Models}
In recent years, Large Language Models (LLMs) like the GPT \citep{floridi2020gpt} family have received more attention due to their performance and potential applications, especially the ChatGPT\citep{radford2019language} and GPT-4\citep{DBLP:journals/corr/abs-2303-08774}. Some variants of those models have been used in many scientific domains, such as BioGPT \citep{luo2022biogpt}, DrugGPT \citep{li2023druggpt}, and MolReGPT \citep{li2023empowering}, which have been adapted for molecular science tasks. Additionally, models like LLaMA and T5 have inspired a variety of variants and as the language model in multi-modal models.

Multi-modal models have been a primary focus on image captioning task \citep{stefanini2022show}. Modality-adaptive Learning models such as BEIT-3 \citep{wang2022image} and KOSMOS-1 \citep{huang2023language} adopt the Mixture of Experts (MoE) strategy \citep{bao2022vlmo} and the MAGNETO transformer \citep{wang2022foundation}, enhancing learning across different modalities using specialized encoders and shared self-attention modules. In contrast, Multi-modal agents like Flamingo \citep{alayrac2022flamingo} and Gato \citep{reed2022generalist} are crafted for real-world applications, leveraging multi-task and reinforcement learning to interpret and act within intricate environments. Additionally, models such as Visual ChatGPT \citep{wu2023visual} employ the Chain-of-Thoughts (CoT) strategy \citep{wei2022chain} approach, merging visual encoders with ChatGPT architectures for enriched, visually-informed dialogues. 
Lastly, the Cross-modal Learning paradigm, as seen in models like MiniGPT-4 \citep{zhu2023minigpt}, emphasizes modality fusion and alignment using techniques like the Q-Former \added{in BLIP2} \citep{li2023blip} or adapter, promoting the modality integration.

In addition, text-image generation models like DALLE-2 \citep{ramesh2022hierarchical} and UniDiffuser \citep{bao2023one} employ diffusion models \citep{yang2022diffusion} to generate images from text. These models have potential applications across art, design, and scientific visualization. Moreover, diffusion models are also used in molecular research for molecule structure graph generation tasks.

In multi-modal tasks, image-captioning and text-image generation stand out as significant directions. These models integrate information across different modalities, highlighting the potential of multi-modal learning in comprehensively understanding the connections in modalities.

\subsection{Multi-modal Model in Molecular Science}
In molecular science, multi-modal models focus on molecule-caption translation tasks and utilize multi-modal representations for downstream tasks like molecular property and chemical reaction prediction.
 
\textbf{Molecule-Caption Translation}, in this task, our model can learn a shared semantic space from a dataset of molecules paired with their text descriptions. Like Text2Mol \citep{edwards2021text2mol}, uses natural language descriptions to retrieve molecules. Moreover, models including KV-PLM \citep{zeng2022deep} and MolT5 \citep{edwards2022translation} significantly contribute to this area. KV-PLM builds a machine reading system, pre-trained on the domain-specific corpus, linking molecules and biomedical text. MolT5 is a self-supervised framework enhancing molecule-caption translation tasks. In addition, MoleculeSTM and MoMu bridge molecular graphs and text data through contrastive learning. MolReGPT applies a retrieval-based paradigm for molecule-caption translation, leveraging LLMs like ChatGPT without fine-tuning.
These models, transforming between SMILES expressions and text descriptions, have the potential for further advancement, with possibilities including better modality alignment, fusion strategies, and fine-tuning methods.

\textbf{Molecule Image Captioning}, rule-based models such as MolVec2 \citep{peryea2019molvec} and OSRA \citep{filippov2009optical}, along with machine learning-based ones like DECIMER \citep{rajan2021decimer} and SwinOCSR. These models use image encoders like Vision Transformer \citep{dosovitskiy2020image} or ResNet \citep{he2016deep} and process image features using recurrent neural networks (RNNs) or transformers \citep{vaswani2017attention} to decode into SMILES strings.
Unlike the general domain, their results mainly focus on transforming images into SMILES strings.
 
\textbf{Molecular Property Prediction}, GNN-based models such as GraphCL \citep{wang2022molecular} and GraphMAE \citep{hou2022graphmae} are used, leveraging contrastive learning and self-supervised graph autoencoders respectively. Models like Uni-mol \citep{zhou2023uni} and GraphMVP \citep{liu2021pre} process 3D graph data effectively, while MoMu and MoleculeSTM combine SMILES and graph representations for downstream tasks.

\added{
The current challenge lies in enhancing modal fusion and merging data from various sources like text, graphs, and images. While these tasks demonstrate the vast potential of multi-modal learning, there remains significant scope for improvement, particularly in capturing the intricate relationships between different data sources. 
To address these challenges, our GIT-Mol model takes inspiration from the BLIP2's Q-Former approach, centering around text modalities while expanding the range of modal adaptability. For handling image modality data, in addition to employing the Swin Transformer, we have integrated contrastive learning strategies into our training regime.
Moreover, our dataset features more complex molecular structures compared to existing models. In contrast to MoMu's graph encoder, GIT-Mol not only employs contrastive learning but also combines it with cross-attention and modal fusion methods.}

\added{
GIT-Mol is adept at mapping multi-modal tensors of varying lengths into a fixed-length unified latent space using the GIT-Former. 
Our model aligns at the entity level of molecules, providing a significant degree of flexibility. 
This adaptability ensures that our model remains versatile and applicable even in different scenarios, such as when replacing images or graphs with other modalities. 
These enhancements enable our model to more effectively capture and integrate the nuanced interplay between various modalities, thereby addressing the critical challenges in the field and pushing the boundaries of molecular science research.
}

\section{Methodology}
\label{sec:Methodology}
\subsection{Overview}
In this work, we present GIT-Mol, a multi-modal large language model for molecular science.
\added{
As shown in Figure \ref{fig:Conceptual-GIT-MOL}, molecular information is categorized into internal and external forms in our research. Internal information, including molecular SMILES and graph data, focuses on the intrinsic structure and rules of molecules. 
External information comprises molecular images and molecule captions, offering a more interpretable perspective. Furthermore, GIT-Mol can fuse data from diverse sources and present a comprehensive view of molecules.}
Specifically, we propose GIT-Former, a novel module capable of aligning all modalities into a unified latent space. It is designed to incorporate various molecular data types, including graphs, images, and text.
In the pre-training phase, the model employs cross-attention and contrastive learning to align different modalities, enriching our understanding of the molecular data. Each fusion strategy within GIT-Former operates at the molecule level, ensuring optimal flexibility and scalability. 
During the fine-tuning stage, the process is guided by prompt learning that adapts to various tasks. Furthermore, these multi-modal representations can be directly processed through MLP to execute molecular properties prediction.

\subsection{Data and Preprocessing Strategy}
\textbf{Data Modalities}: The diversity of data modalities allows the model to learn and understand complex relationships across different modalities, which improves the performance in molecular property prediction and generation tasks. Molecular information is categorized as internal and external information, as shown in Figure 1. Internal data like SMILES strings and structural graphs are essential for predicting molecular properties and features. 
The SMILES strings provide a textual representation of molecular structures, concisely encoding vital connectivity and stereochemistry details. Furthermore, molecular structured graphs offer a topological view of molecules in two-dimensional space, where atoms are nodes and bonds are edges. 
In contrast, external data, including text descriptions and molecule images, are user-friendly and easy to interpret.
Molecular captions present textual descriptions that shed light on molecules' distinct characteristics and properties, offering a natural language context for the model. Furthermore, molecular images visually showcase atomic structures and bonding schemes, providing intuitive input for our model's molecular analysis.

\textbf{\added{Dataset}}: We collected approximately 4.8 million chemical compounds from the PubChem \citep{kim2019pubchem} database, providing a robust training dataset for our model. This dataset contains molecular images for image captioning tasks and serves as a rich resource for self-supervised learning with SMILES and molecular graph representations.
In addition, we use the standard ChEBI-20 dataset \citep{edwards2021text2mol}, consisting of 33,010 molecule-description pairs, for fine-tuning and evaluation. While the molecule captions in some databases are less complex than those in ChEBI-20, it is crucial to have concise and accurate descriptions of molecular characteristics. 
\added{
We constructed a dataset from ChEBI \citep{hastings2016chebi} and PubChem. PubChem has over 320,000 molecule-description pairs, but many of them are too brief or contain non-informative content.
To ensure high-quality captions, we analyzed the length distribution of captions in the CHEBI-20 dataset. We established a criterion for selecting captions that are longer than 96 characters and excluded those without any relevant information on their properties or functional groups.}

\textbf{Data Preprocessing}: To ensure optimal model training and evaluation, we undertake several critical steps to refine raw data. Initially, we focus on data cleaning by rectifying inconsistencies, filling in missing values, or resolving data errors. It guarantees a high-quality data input for the model. 
Further, we remove compounds that feature low-frequency atoms by analyzing atom counts. It ensures the model focuses on patterns from prevalent and pertinent atomic structures. 
Lastly, using the RDKit toolkit \citep{bento2020openx}, we validate and enhance the dataset's graph structures. We refine our collection to include only high-quality and representative compounds. 
By rigorously preprocessing the data, we assure data quality and bolster the model's learning efficiency and performance on targeted tasks.

\subsection{GIT-Former}
GIT-Former is an architecture that can map all modalities into a unified latent space, designed based on the Q-Former architecture in BLIP2 \citep{li2023blip}. 
It leverages the strengths of various encoder and decoder models to suit the specific characteristics of different data modalities, such as SMILES strings, captions, images, and structure graphs.  
This section first introduces the model architecture. Then, it delineates the pre-training and fine-tuning in two stages: (1) the Xmodal-to-language representation learning stage with frozen encoders and (2) the Xmodal-to-language generative learning stage with the decoder. 
The GIT-Former model architecture and training strategy are shown in Figure \ref{fig:Conceptual-GIT-MOL}. \added{In the pre-training phase of GIT-Former, attention mechanisms and contrastive learning are used to align Xmodal data with the target text modal. During the fine-tuning phase, Xmodal data is mapped onto learnable fixed-length queries, effectively aligning diverse modalities into a latent space.}

\textbf{GIT-Former} model serves as a modality mixer to 
 fuse molecular data. Unlike existing Visual Language (VL) models like BLIP2's Q-Former, our approach can address the graph and text modality translation tasks, providing the flexibility needed in bio-molecular multi-modal studies. To better fit our scientific scenario, we replace the BERT \citep{devlin2018bert} model with SciBERT \citep{beltagy2019scibert}, tailored explicitly to scientific text, and enhance it with cross-attention mechanisms for graph and text modalities. As shown in Figure \ref{fig:Conceptual-GIT-MOL}, our model can adaptively adjust its training modules to achieve modality alignment effectively. Moreover, alternatives such as MolT5 can also be utilized as embedding and self-attention layers, providing flexibility to adapt to different molecular scenarios.

\textbf{Encoder and Decoder}: MolT5 is an advanced language model to translate molecule and molecular textual descriptions. We adopt it to serve as our text encoder and decoder. For image encoding, we adopt the Swin Transformer \citep{liu2021swin} from SwinOCSR, and for graph encoding, we select the GIN model from the pre-trained MoMu model. This multi-encoder and decoder setup equips our model with the flexibility to adapt to the demands of each data modality, enhancing the model's performance and efficacy.

\textbf{Cross-Attention Mechanism} is central to GIT-Former. This module facilitates the alignment of image and graph modalities to text modalities, which encompass molecular captions and SMILES strings. This mechanism can be mathematically formulated as a multi-head attention operation, which captures inter-modal data relationships:

\begin{equation}
\text{MultiHead}(Q_{Tt}, K, V) = \text{Concat}(\text{head}_1, \ldots, \text{head}_h)W_O
\end{equation}
\begin{equation}
\text{head}_i = \text{Attention}(Q_{Tt}W_{Qi}, KW_{Ki}, VW_{Vi})
\end{equation}

In the context of GIT-Former, \( {Q} \) represents queries, \( Q_{Tt} \) represents the queries from the target text modality (caption or SMILES string), \( {K} \) stands for keys, \( {V} \)  denotes values, and \( {W} \)  refers to the learned weight matrices to project the input into appropriate spaces.
For source modalities in equation (2), \( {i} \) represent the graph, image, and source text modality. For each source modality, GIT-Former computes a separate set of cross-attention weights, aligning the \( Tt \) modality to each of the source modalities. This ensures that every detail of the image or graph is precisely reflected in its corresponding textual representation.
The attention mechanism can be visualized as:
\begin{equation}
\text{Attention}(Q_{Tt}, K, V) = \text{softmax}\left(\frac{Q_{Tt}K^T}{\sqrt{d_k}}\right) V
\end{equation}
where \( d_k \) is the dimensionality of the queries and keys.

By leveraging this cross-attention framework, GIT-Former efficiently discerns and encodes intricate relationships and dependencies between modalities. This is shown in Figure 1, emphasizing the alignment between image, graph and source text modalities to \( Tt \) modality.

\subsection{Pre-training Strategy}
Our research involves various training strategies. In the pre-training phase, we use frozen image and graph encoders and SciBERT for text encoding. 
Moreover, the GIT-Former aligns each modality with the target text by self-supervised learning, enhancing molecular translation tasks. 
We use a unique cross-attention method in pre-training to achieve inter-modal and contrastive learning.

\textbf{Xmodal-Text Matching (XTM)} aims to align different modal representations with corresponding text. It is a binary classification task that determines if a set of cross-modal texts matches. 
A bi-directional self-attention mask facilitates interaction between learnable queries and modality embeddings.
\added{The learnable query is an initialized, fixed-length tensor of zeros. The embeddings from different modalities are mapped onto this fixed-length tensor through cross-attention mechanisms. 
This process results in the generation of query embeddings enriched with multi-modal information.
}
These \added{query embeddings} are processed through a linear classifier to get a logit, and the average across queries gives the final matching score.
For the different modalities (X-source-modal and text-target-modal), we convert the inputs into embeddings ($E_x$ and $E_t$) and then pair them with matched and mismatched samples, creating a contrast for the model to recognize the correct combinations. Self-attention and cross-attention mechanisms are employed to understand relations within and across modalities, resulting in a unified embedding that contains multi-modal information. This process can be represented with $f_{att}$ as the function for both attention mechanisms.
\begin{equation}
E_{\text{fused}} = f_{\text{att}}(\text{concat}([E_x, E_t]))
\end{equation}

Lastly, the fused embeddings \(E_{\text{fused}}\)  are processed through a linear layer with a weight of $W$, yielding a predictive score (logit). The cross-entropy loss between the logit and actual labels (1 for match, 0 for mismatch) is computed. This loss is a metric for the model's performance on the binary classification task, and its optimization helps the model effectively match information across different modalities.
\begin{equation}
\text{loss}_{\text{xtm}} = \text{cross\_entropy}(W \cdot E_{\text{fused}}, y)
\end{equation}

\textbf{Xmodal-Text Contrastive Learning (XTC)} aligns different information types with corresponding text representations. This technique contrasts the similarity of matched cross-modal text against mismatches. The representation from the task modality is aligned with the text, and the one with the highest similarity is selected as the cross-modal data pair. In the XTC approach, we first employ the GIT-Former to attain the learned representation of the Xmodal input. We then extract the Xmodal features $E_x$ from this representation. Subsequently, we compute the mutual information $I(E_x, E_t)$, individually for Xmodal and text representations $E_t$. Finally, we use cross-entropy to calculate the loss function $L_{xtc}$. 
The operation \(\cdot\) denotes the dot product between the two embeddings. With this in place, the mutual information is determined as:
\begin{equation}
I(E_x, E_t) = E_x \cdot E_t
\end{equation}

Finally, the cross-entropy loss \(L_{xtc}\) is computed as:
\begin{equation}
L_{xtc} = \text{cross\_entropy}(I(E_x, E_t), y)
\end{equation}

\subsection{Fine-tuning Strategy}

\added{
During the fine-tuning phase, we classify our tasks into two main categories: Modality Translation and Molecular Property Prediction. Within the Modality Translation task, as illustrated in Figure \ref{fig:Conceptual-GIT-MOL}, our focus areas include molecule captioning, molecule generation, and molecular image recognition. To enhance the flexibility of this task, we incorporate our prompt manager, which enables an any-to-language training mechanism, facilitating a more versatile modality translation process.
In the Molecular Property Prediction task, we fine-tune our model using labeled SMILES and graph data from MoleculeNet \citep{wu2018moleculenet}, specifically classification tasks. 
}

\textbf{Prompt Tuning for Modality Translation}: As shown in Figure 1, our model encodes data from each modality into embeddings, and then the GIT-Former maps embeddings into a unified latent space. This allows seamless interaction and translation among different modalities within a common representation. Moreover, the framework supports translating information across any modality to the language modality, providing a flexible setup for managing tasks such as prompt management, data loading, and model training. This is achieved by designing various task types and corresponding prompts according to the specific modality.

\added{The any-to-language molecular translation strategy fundamentally revolves around a text-centric pretraining and finetuning approach.
It can be implemented through a combination of prompts, GIT-Former, and Large Language Models (LLMs) to accomplish any-to-language tasks.
The strength of this strategy lies in its flexibility in modal transformation and adaptability to various tasks.
}
\added{We test directive and guiding prompts for each task.}
Finally, we employ the following prompt for the SMILES string generation and recognition task: "Given the provided {$inputs$}, generate the corresponding SMILES string." Once this prompt is fused with the GIT-Former embeddings, the combined representation is channeled to the MolT5 decoder. Here, the training objective is to align the input data (a caption or an image) with its pertinent SMILES data. This design ensures that the model remains sensitive to the nuanced differences between modalities while still producing reliable SMILES representations, bridging the gap between visual or structural information and text.

In the molecule captioning task, our prompt choice is: "This is the {$inputs$} of the molecule, and the corresponding caption is: " Similar to the previously outlined procedure, upon amalgamation of this prompt with the GIT-Former embeddings, the aggregate is routed to the MolT5 decoder. The overarching training goal in this scenario revolves around ensuring that the input data (SMILES string or graph) aligns impeccably with its associated caption data. This task reinforces the model's ability to understand molecular structures from various perspectives. It enriches its competency in generating descriptive text, demonstrating the model's broad applicability in multi-modal molecular informatics.

\textbf{Fine-tuning for Property Prediction Tasks}: We use MolT5 as embedding and self-attention layers to process SMILES strings and employ contrastive learning to pre-train SMILES string and graph embeddings. Our approach emphasizes utilizing multi-modal data integrating modality representations through attention mechanisms.
After obtaining this enriched embedding, it is passed through a Multi-Layer Perceptron (MLP), which outputs the predicted molecular properties.

Language models have inherent strengths in understanding context and relationships, our model is particularly apt for benchmark tasks related to bio-activity and toxicity classifications. The model's ability to process and relate complex molecular data makes it well-suited for these specific prediction challenges.

\subsection{The Essentiality of Architecture and Strategy}

In molecular science, molecular data have multiple representation formats, such as molecular images, structural graphs, SMILES strings, and physical properties. Most existing models bridge two modalities into a unified space, like images with SMILES strings or graph structures with captions. In summary, accommodating an expanded set of modalities remains challenging for current models.

\textbf{Architecture}: GIT-Former provides a dynamic and scalable solution tailored for multi-modal molecular data. The prowess of GIT-Former emanates from its cross-attention mechanism, allowing it to integrate and adapt to new modalities within molecular representations effortlessly.
\added{
The key feature of the GIT-Former is its ability to map multi-modal tensors of varying lengths into a fixed-length unified latent space. This mapping allows the model to align data at the molecular entity level, offering significant flexibility.
Such adaptability ensures that GIT-Mol remains versatile and applicable in various scenarios, including those involving modalities beyond images and graphs.}

\textbf{Training strategies}: The choice of XTC and XTM as training strategies, both strategies operate at the individual vector level, providing flexibility in model training. XTC is crucial for distinguishing between positive and negative molecular samples, especially since minor molecular changes can lead to significant shifts in bioactivity or drug interactions. Meanwhile, XTM ensures that each molecular representation aligns precisely with its corresponding text or description.

\textbf{Multi-modal Representation}: GIT-Former can extract information from each modality and ensure that this information effectively integrates within a unified representation space. It ensures a deeper understanding of the nuances within each modality. Moreover, when paired with prompt learning and the pre-trained state-of-the-art MolT5 decoder, GIT-Mol effectively leverages this information in specific tasks, producing the corresponding molecular data.

\section{Results}
\label{sec:Results}
In our experiments, we delve into crucial tasks related to molecular multi-modal translation, including molecule captioning (Sec \ref{sec: Molecule Captioning}), text-based de novo molecule generation (Sec \ref{sec: Molecule Generation}), and molecular image recognition (Sec \ref{sec: Molecule Image Captioning}). 
Additionally, we utilize the processed embeddings for molecular property prediction tasks for the molecular multi-modal representation (Sec \ref{sec: Molecular Property Prediction}). Additionally, we have instituted ablation studies to ascertain the contributions of multi-modal data, cross-attention, and prompt learning to the overall performance.
To evaluate our model's efficacy, we compare embeddings pre-model and post-model processing, aiming to understand the model's impact on molecular data (Sec \ref{sec: Embeddings Visualization}). Finally, we detail the training configurations and hyperparameters employed in our experiments (Sec \ref{sec: Training Settings}).

\subsection{Molecule Captioning}
\label{sec: Molecule Captioning}

\textbf{Experimental Setup}: Molecule captioning primarily focuses on generating textual descriptions of molecules based on their inherent information, such as their SMILES strings, graph representations, and other molecular features. To evaluate the performance of our model, we conduct experiments using a set of established Natural Language Processing (NLP) metrics. These metrics include BLEU, which measures the n-gram overlap between generated and reference sentences; ROUGE, which emphasizes the overlap of various n-grams, word sequences, and word pairs; and METEOR, a comprehensive metric that takes into account precision, recall, synonymy, stemming, and word order. These collectively assess the quality and relevance of the generated text in molecule captioning. \added{Within the framework of our GIT-Mol model, SciBERT and MolT5 serve as language model components, and then we choose SciBERT and MolT5 as the baselines for our model. This selection is particularly pertinent in our Molecule Generation experiments.}

\textbf{Molecule Caption Dataset}: Our team procured a substantial dataset of 320,000 molecules accompanied by detailed descriptions from the PubChem Record Description category. We enhanced this dataset by conducting a thorough crawl of the corresponding captions (up to a maximum of 3) from each molecule page. We augmented our dataset by combining it with the ChEBI-20 Dataset, thus forming a highly comprehensive and informative multi-modal caption dataset.

We further carried out data cleaning to remove "invalid" descriptions in the captions, such as those that merely specify the source of the compound without any pertinent information about its properties or characteristics. However, we opted to retain succinct descriptions that provide essential information about the compound. This enriched dataset consists of approximately 90,000 data points, incorporating multi-modal data such as images of the molecules, SMILES strings, and captions.

\begin{table*}[htbp]
	\centering 
	\begin{tabular}{c|c|c|c|c|c|c}
	\hline
	\textbf{Model} & \textbf{BLEU-2} & \textbf{BLEU-4} & \textbf{ROUGH-1} & \textbf{ROUGH-2} & \textbf{ROUGH-L} & \textbf{METEOR}\\
		\hline
            \hline
            SciBERT & 0.184 & 0.113 & 0.412 & 0.327 & 0.397& 0.367\\	
		MolT5-base & 0.316 & 0.247 & 0.572 & 0.480 & 0.545 & 0.529\\	\hline
            GIT-Mol(SMILES) & 0.264  & 0.176 & 0.477 & 0.374 & 0.451 & 0.430\\
		GIT-Mol(Graph) & 0.290  & 0.210 & 0.540 & 0.445 & 0.512 & 0.491\\
         \hline
        GIT-Mol(XTM)  & 0.264  & 0.187 & 0.521 & 0.421& 0.494 & 0.471\\
		\hline
         GIT-Mol $op$ & 0.312  & 0.237 & 0.556 & 0.468 & 0.535 & 0.525\\
            GIT-Mol  & \textbf{0.352}  & \textbf{0.263} & \textbf{0.575} & \textbf{0.485}& \textbf{0.560} & \textbf{0.533}\\
		\hline
	\end{tabular}
    \caption{\textbf{Molecule captioning results}. In the subsequent experimental results, $op$ signifies models that do not utilize prompt learning, XTM represents models that only utilize XTM pre-training strategy.}
    \label{tabMolecule captioning results}  
\end{table*}

\begin{figure*}[htbp]
\centering
\includegraphics[width=1.0\textwidth]{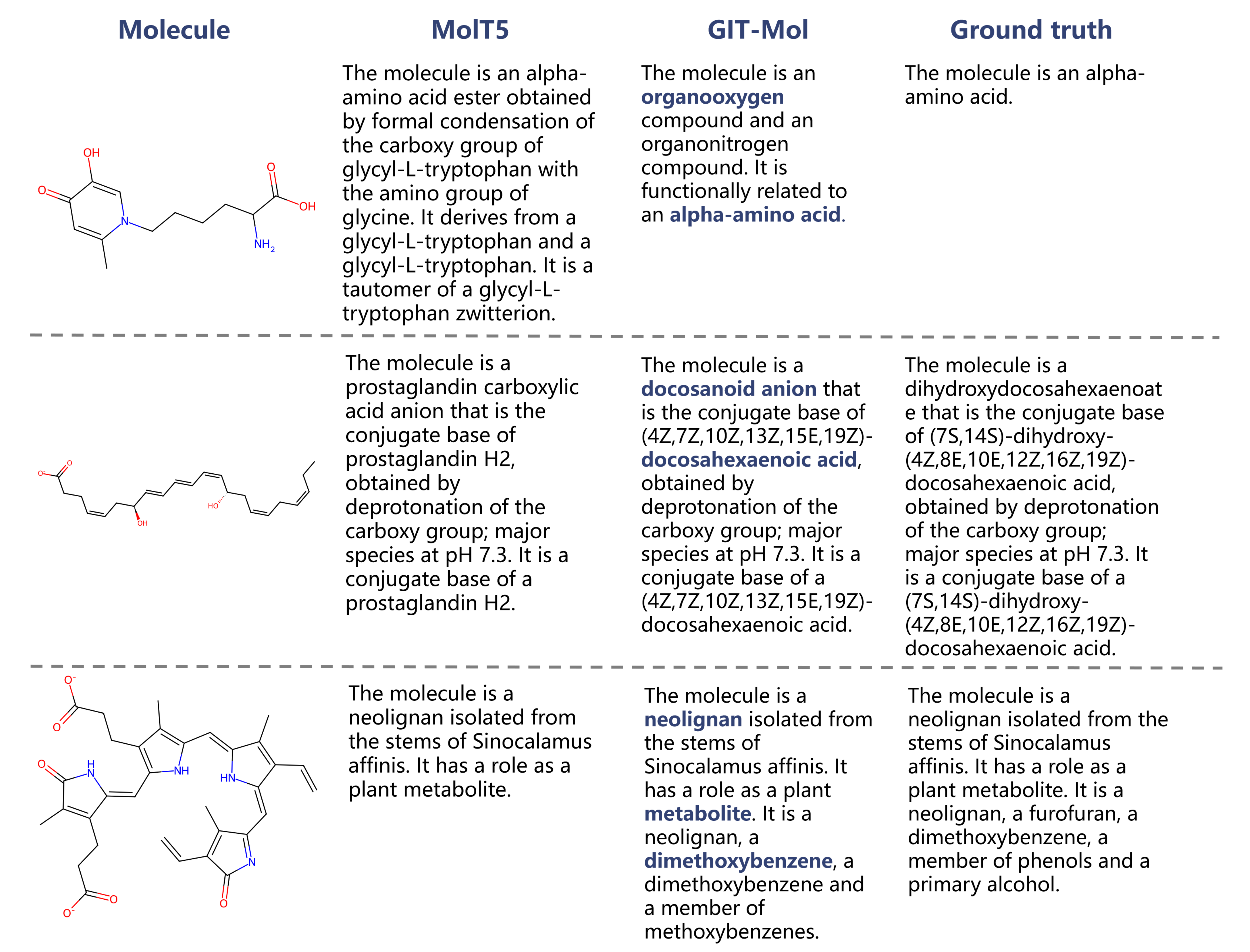}
\caption{\textbf{Study case of Molecule Caption}. The GIT-Mol model exhibits precise chemical characterization, aligning closely with ground truth information.}
\label{fig: Study case1}
\end{figure*}

\textbf{Results and Observations}: Our model demonstrates superior performance in generating high-quality, relevant molecule captions. Detailed results are shown in Table \ref{tabMolecule captioning results}. This experiment underscores the potential of multi-modality and suggests areas for further optimization and exploration.The GIT-Mol(graph) model delivers better performance than GIT-Mol(SMILES) across all metrics. 
It implies that graph representations might capture molecular structures more effectively than SMILES strings. 
However, neither of them outperforms the multi-modal model.
\added{In the ablation study, the employment of this multi-modal model resulted in an impressive 10\%-15\% improvement over single-modality models.
It suggests that each modality brings complementary information to the task.
As for pre-training strategies, the model pre-trained only with XTM shows room for enhancement, underlining the effectiveness of the XTC strategy in boosting overall performance.}

\textbf{Case Studies}: We select a set of molecules, feed them to our model, and let it generate captions based on the learned representations. The results are shown in Figure \ref{fig: Study case1}. \added{For each molecule, we present the ground truth, the captions generated by our model and the baseline model.}

\subsection{Text-Based de novo Molecule Generation}
\label{sec: Molecule Generation}

\textbf{Experimental Setup}: Text-Based de novo Molecule Generation aims to produce molecular SMILES representations based on provided textual captions of molecules. To gauge the accuracy and validity of the molecules generated by our model, we employ a mix of cheminformatics and NLP metrics:
\begin{itemize}
    \item \textbf{Fingerprint Tanimoto Similarity (FTS)}: We utilize this to measure the chemical similarity between the ground-truth molecules and the generated ones, considering multiple fingerprint types: MACCS, RDK, and Morgan fingerprints. FTS provides a decimal value between 0 (no similarity) and 1 (perfect similarity), offering a quantifiable measure of chemical resemblance.
    \item \textbf{BLEU}: Used primarily in NLP tasks, BLEU scores evaluate the overlap of n-grams between generated and reference SMILES strings, signifying the closeness of the generated SMILES strings to the actual ones.
    \item \textbf{Exact Match}: It simply checks if the generated SMILES string matches the reference SMILES string. An exact match score directly measures the model's precision in producing accurate molecular representations.
    \item \textbf{Levenshtein distance}: This metric calculates the minimum number of single-character edits (i.e., insertions, deletions, or substitutions) needed to change one SMILES string into another. A smaller distance indicates a closer similarity between the generated and reference SMILES string.
    \item \textbf{Validity}: We employ RDKit, a cheminformatics software, to check the chemical validity of the generated SMILES string. By reporting the percentage of valid molecules, we ensure that our model does not just produce semantically accurate SMILES strings but also chemically feasible ones.
\end{itemize}
To stay consistent with our methodology in the molecule captioning task, we employ MolT5-base as both the decoder and the baseline for comparison in our experimental setup.

\textbf{Molecule Generation Dataset}: We use the same ChEBI-20 Dataset as MolT5 for fine-tuning our model on the molecule generation task. It contains various molecular structures represented in SMILES notation, providing a rich resource for training models to generate novel molecules. 

\textbf{Results and Observations}: Our model exhibits impressive results in generating chemically valid molecules that closely match the ground-truth molecules regarding chemical and structural characteristics. 
The results are shown in Table \ref{tab: Molecule generation results}. Compared to the MolT5-base(caption) model, GIT-Mol(caption) slightly outperforms the MACCS FTS metric and presents closely competitive results on the other two molecular similarity measures, RDK FTS and Morgan FTS.
Notably, the performance of GIT-Mol(caption) on the “Validity” metric reaches an impressive 0.928, significantly higher than the other two models. 
This indicates that GIT-Mol can generate a higher proportion of valid molecules while still preserving high molecular similarity. 
\added{In the ablation study, the model pre-trained only with XTM displays a higher propensity for overfitting, with a lack of answer diversity and a tendency to generate homogenous molecular structures.
Additionally, we demonstrate the necessity of prompt learning in influencing the performance of this task.} 
We provide some examples of molecule generation based on text descriptions and detailed analysis. 

\begin{table*}[htbp]
	\centering
	\begin{tabular}{c|c|c|c|c|c|c|c}
        \hline
	\textbf{Model} & \textbf{BLEU} & \textbf{Exact} & \textbf{Levenshtein} & \textbf{MACCS FTS} & \textbf{RDK FTS} & \textbf{Morgan FTS} & \textbf{Validity}\\
		\hline
            \hline
         SciBERT & 0.459 & 0.005 & 55.459 & 0.499 & 0.344 & 0.254 & 0.915\\	\hline
		MolT5-base& \textbf{0.769} & \textbf{0.081} & \textbf{24.458} & 0.721 & \textbf{0.588} & \textbf{0.529} & 0.772\\	\hline
        GIT-Mol(XTM) & 0.245 & 0.0 & 82.633 & 0.322 & 0.196 & 0.087 & 1.0\\	
        \hline
            GIT-Mol $op$ & 0.721  & 0.041 & 30.41 & 0.705 & 0.477 & 0.453 & 0.812\\
		GIT-Mol & 0.756  & 0.051 & 26.315 & \textbf{0.738} & 0.582 & 0.519 & \textbf{0.928}\\
		\hline
	\end{tabular}
    \caption{\textbf{Molecule generation results}. Our model performs similarly to MolT5-base for generating molecular SMILES strings but excels in ensuring molecule validity. With a molecule validity of \textbf{92.8\%}, our model surpasses MolT5-base's validity of 77.2\% by over \textbf{20\%}.}
    \label{tab: Molecule generation results}
\end{table*}

\begin{figure*}[htbp]
\centering
\includegraphics[width=1.0\textwidth]{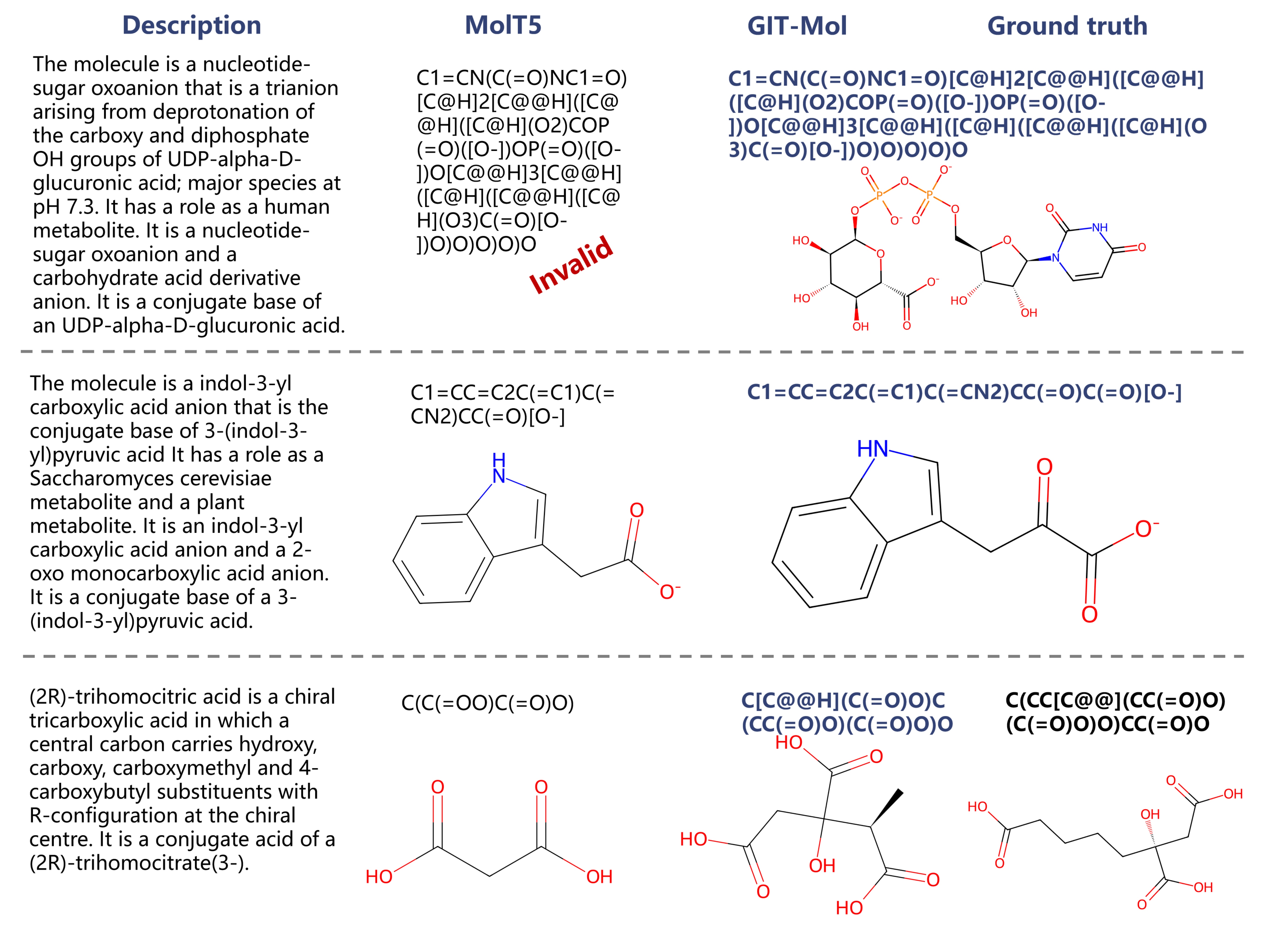}
\caption{\textbf{Study case of Molecule Generation.} GIT-Mol stands out for its ability to generate valid molecules. \added{Besides, even if not identical to the ground truth, it still faithfully adhere to the features described in the textual captions.}}
\label{fig: Study case2}
\end{figure*}

\textbf{Case Studies}: In Figure \ref{fig: Study case2}, we showcase the outcomes for every molecule instance. Our model, the baseline model, and GPT4 all generate SMILES strings, which are compared to the ground truth. This comparison allows for a visual evaluation of the quality and relevance of the generated molecules, offering a better understanding of our model's effectiveness.

\subsection{Molecule Image Recognition}
\label{sec: Molecule Image Captioning}

\textbf{Experimental Setup}:  Molecular Image Recognition focuses on converting 2D molecular images into their corresponding SMILES representations. We employ the identical set of metrics as those used in "Text-Based de novo Molecule Generation." For a fair and rigorous assessment, we juxtapose our model with SwinOCSR, an acknowledged efficient model in molecular image captioning.

\textbf{Molecule Image Dataset}: We employ the same dataset as the Molecule Caption Dataset utilized in the Molecule Captioning Task. The distinction lies in the training data composition, where the pairings are between SMILES strings and their corresponding molecular images rather than textual descriptions.

\textbf{Results and Observations}: The results are presented in Table \ref{tab: Molecule image recognition}. The empirical outcomes indicate that our model, whether employing prompt learning or not, the GIT-Mol model outperforms the SwinOCSR model across all evaluated metrics. It emphasizes the efficacy of the GIT-Mol architecture in the molecule image recognition task. \added{In the ablation study, similar to findings in the molecule generation experiment, the model exclusively using the XTM pre-training strategy not only overfitted more easily but also converged more slowly, resulting in the generation of uniform molecular structures. 
Our model demonstrates the model's extensibility and effectiveness of prompt learning.}
 \begin{table*}[htbp]
	\centering
	\begin{tabular}{c|c|c|c|c|c|c|c}
	\hline

	\textbf{Model} & \textbf{BLEU} & \textbf{Exact} & \textbf{Levenshtein} & \textbf{MACCS FTS} & \textbf{RDK FTS} & \textbf{Morgan FTS} & \textbf{Validity}\\
		\hline
            \hline
		SwinOCSR & 0.892 & 0.376 & 9.157 & 0.945 & 0.872 & 0.846 & 0.827\\	\hline
            GIT-Mol(XTM) & 0.347 & 0 & 62.075 & 0.423 & 0.289 & 0.137 & 1 \\	\hline
            GIT-Mol $op$ & 0.913 & 0.405 & 8.675 & 0.957 & 0.885 & 0.878 & 0.850\\	
		GIT-Mol & \textbf{0.924} & \textbf{0.461} & \textbf{6.575} & \textbf{0.962} & \textbf{0.906} & \textbf{0.894} & \textbf{0.899}\\
		\hline
	\end{tabular}
      \caption{\textbf{Molecule image recognition results}. \added{The results suggest that prompt learning aids in producing more accurate SMILES strings from molecule images.}}
      \label{tab: Molecule image recognition}
\end{table*}

\subsection{Molecular Property Prediction}
\label{sec: Molecular Property Prediction}

\textbf{Experimental Setup}: In our study, we use six essential classification datasets related to molecular biological activity from MoleculeNet \citep{wu2018moleculenet} , including Tox21, ToxCast, Sider, ClinTox, Bace, and BBBP. These datasets contribute to understanding a molecular properties and effects. We ensure the model's fairness and robustness by using a scaffold-based splitting approach for structuring training, validation, and test sets. This leads to diverse molecule evaluations. The pre-trained model on PubChem processes the datasets into molecule embeddings, and we optimize using the Binary Cross-Entropy (BCE) loss function.

\textbf{MoleculeNet for Molecular Property Prediction.} In our study, we chose a few classification task datasets from MoleculeNet as our fine-tuning datasets. These datasets include Tox21, ToxCast, Sider, ClinTox, BBBP, and Bace, each covering different molecular property prediction tasks.
\begin{itemize}
\item \textbf{Tox21}: A collaborative project to identify potential toxins,  containing toxicity data for approximately 7800 compounds across 12 different pathways.
\item \textbf{ToxCast}: Contains biological activity data for about 8600 environmental compounds from over 600 in vitro experiments, used to assess potential toxicity.
\item \textbf{Sider}: A database of side effect information for around 1400 drugs, used for predicting potential drug side effects.
\item \textbf{ClinTox}: Includes toxicity information for known drugs and clinical toxicity predictions for unapproved compounds, primarily used to predict clinical toxicity.
\item \textbf{BBBP}: Utilized to predict whether compounds can penetrate the blood-brain barrier, aiding in understanding brain delivery capacity.
\item \textbf{Bace}: Contains inhibitory activity information on beta-secretase 1 (BACE1) for 1513 compounds, useful in identifying potential candidate drugs for disease treatment.
\end{itemize}

\textbf{Results and Observations}: Our model performs commendably in predicting molecular properties, achieving high AUC scores across multiple runs. 
The results of our model and baselines are shown in Table \ref{tab: MoleculeNet results}.
\added{We demonstrate the effectiveness of multi-modal data combining graph and SMILES for molecular representation, which shows an average performance improvement of 5\%-10\%.
}
\begin{table*}[htbp]
	\centering
	\begin{tabular}{c|c|c|c|c|c|c|c}
	\hline

	\textbf{Dataset} & \textbf{Tox21 ↑} & \textbf{ToxCast ↑} & \textbf{Sider ↑} & \textbf{ClinTox ↑} & \textbf{BBBP ↑}& \textbf{Bace ↑}&\textbf{Avg}\\
        \textbf{Molecules} & 7831 & 8575 & 1427 & 1478 & 2039& 1513&--\\
        \textbf{Task} & 12 & 617 & 27 & 2 & 1& 1&--\\
		\hline
          \hline
        KV-PLM & 72.1$\pm$1.0 & 55.0$\pm$1.7 & 59.8$\pm$0.6 & - & 70.5$\pm$0.5 & 78.5$\pm$2.7 
        & 67.20  \\
        GraphCL & 75.1$\pm$0.7 & 63.0$\pm$0.4 & 59.8$\pm$1.3 & 77.5$\pm$3.8 & 67.8$\pm$2.4 & 74.6$\pm$2.1
        & 69.64 \\
        GraphMVP & 74.9$\pm$0.5 & 63.1$\pm$0.2 & 60.2$\pm$0.13 & 79.1$\pm$2.8 & 70.8$\pm$0.5 & 79.3$\pm$1.5
        & 71.23 \\
        MoMu & 75.6$\pm$0.3 & 63.4$\pm$0.5 & 60.5$\pm$0.9 & 79.9$\pm$4.1 & 70.5$\pm$2.0 & 76.7$\pm$2.1
        & 71.1 \\
        Mole-BERT & \textbf{76.8$\pm$0.5 }& 64.3$\pm$0.2 & 62.8$\pm$1.1 & 78.9$\pm$3.0 
        & 71.9$\pm$1.6 & 80.8$\pm$1.4 &72.58 \\
        \hline
        GIT-Mol(SMILES) & 73.9$\pm$0.7 & 62.1$\pm$0.4 & 60.1$\pm$1.1 & 83.5$\pm$3.1 & 71.9$\pm$1.8 & 68.4$\pm$1.7 & 70.0 \\
        GIT-Mol(Graph) & 75.4$\pm$0.5 & 65.3$\pm$0.7 & 58.2$\pm$0.9 & 78.9$\pm$2.5 & 71.1$\pm$1.5 & 
        65.8$\pm$1.8 & 69.1 \\
        GIT-Mol(G+S) & 75.9$\pm$0.5 &\textbf{66.8$\pm$0.5} & \textbf{63.4$\pm$0.8} & 
        \textbf{88.3$\pm$1.2}& 
        \textbf{73.9$\pm$0.6} & 
        \textbf{81.08$\pm$1.5} & 
        \textbf{74.90} \\
		\hline
	\end{tabular}
        \caption{ \textbf{Results for molecular property prediction (classification)}. The combined use of \textbf{SMILES} and \textbf{2D graphs} enhances our multi-modal molecular representation, which outperforms both single-modal models and other multi-modal approaches.}
        \label{tab: MoleculeNet results} 
\end{table*}

\subsection{\added{Embeddings Visualization}}
\label{sec: Embeddings Visualization}
\begin{figure*}[t!]
\centering
\includegraphics[width=0.80\textwidth]{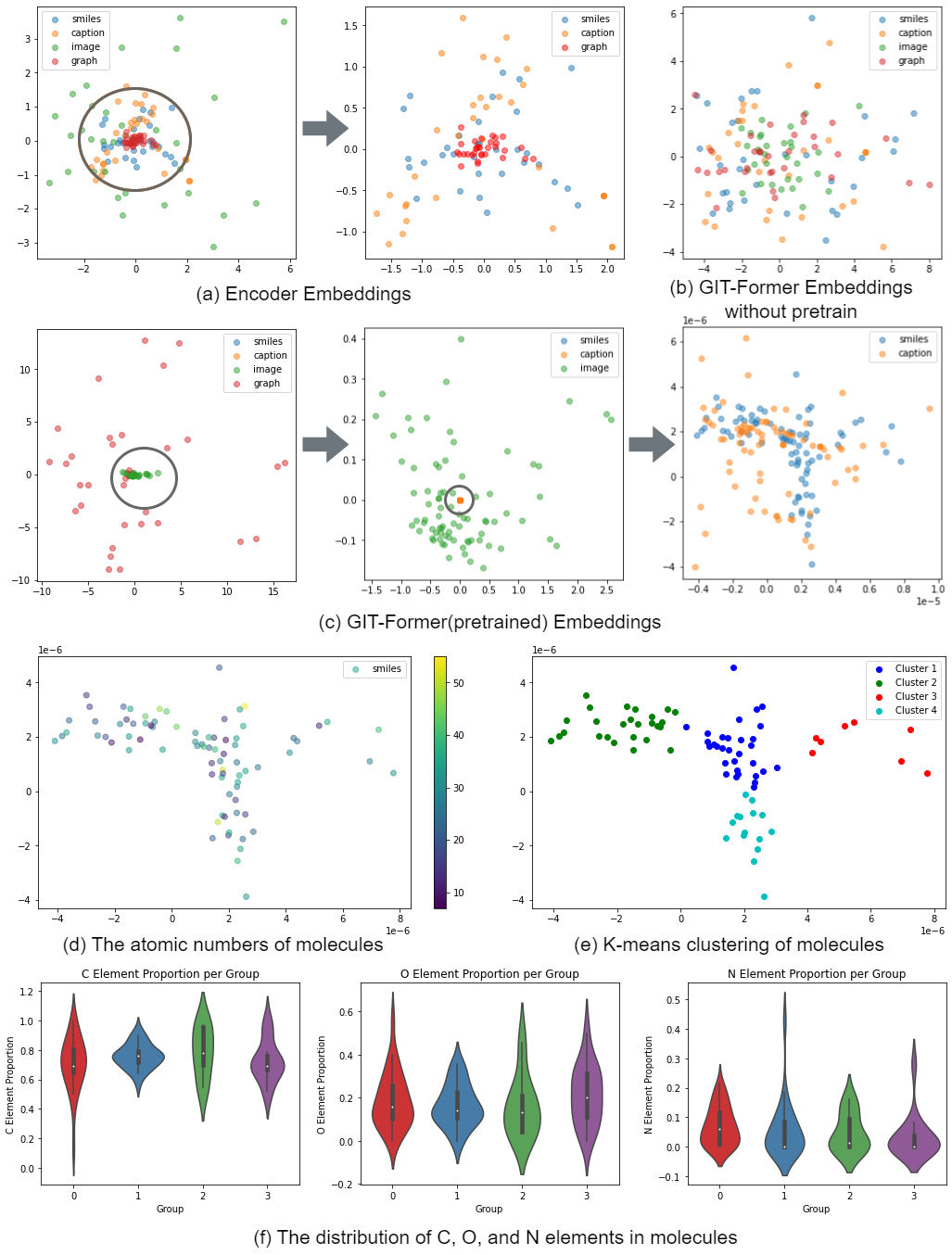}
\caption{\textbf{Embeddings visualization}. (a) Original vector representations from various molecular data modalities. (b) Vector distribution is processed by an untrained GIT-Former, illustrating a tendency towards uniformity. (c) Hierarchical vector distribution processed by a pre-trained GIT-Former showcasing the layered separation of modalities with the outermost layer representing graph embeddings, followed by image embeddings, and innermost containing SMILES strings and captions from the text modality. (d) Distribution of atoms in molecules, with color gradients indicating increasing atom numbers. (e) Results of K-means clustering applied to molecular data. (f) Distribution of C, N, and O atoms across different clusters. The pre-training effects demonstrate GIT-Former's ability to differentiate among modalities, subtypes within a modality, and specific properties within a given data type.}
\label{fig: embedding_vis}
\end{figure*}

\textbf{GIT-Former Processing}:  We experimented on 80 molecules to create vector visualizations, as shown in Figure \ref{fig: embedding_vis}. The first image in Figure 4(a) shows the vector representations created by the encoders for each modality of molecular data. The second image displays the vector representations without the image embeddings. Our analysis of these images reveals that the graph vectors are more concentrated in the original vector representations, while the image representations are more diverse. Additionally, the SMILES and caption representations, both text-based, are not easily distinguishable.
 
\textbf{Pre-training Effects}: In Figure 4(b), we display the vectors processed by an untrained GIT-Former. All vectors appear to move towards a uniform distribution. In contrast, Figure 4(c) shows the vectors processed by a pre-trained GIT-Former. The vector distribution shifts from dispersed to concentrated, with the outermost layer being the graph embeddings, the image embeddings, and the SMILES strings and captions in the text modality. This occurs because our GIT-Former's pre-training strategy involves aligning other modalities to the text modality. The SMILES strings and captions can be further distinguished from the original vector representations.

\textbf{Clustering and Distribution of atoms}: The image in Figure 4(d) displays the distribution of the number of atoms in a molecule. The color changes from dark to light, indicating an increase in the number of atoms. The distribution of the number of atoms is similar when viewed on the same vertical axis. Moving on to Figure 4(e), we conducted K-means clustering on the molecules. We observed the distribution of C, N, and O atoms in Figure 4(f) based on different groups. We can see some differences in the distribution among the different groups. However, by representing different molecules based on these differences, our GIT-Former can differentiate between different modalities, data types within the same modality, and molecular properties within the same data type.

\subsection{Training Settings}
\label{sec: Training Settings}
To achieve optimal performance with our GIT-Former model, we carefully designed both pretraining and fine-tuning stages. This section provides an overview of the training configurations, detailing the encoder settings, hyperparameter values, and the specific adjustments made for different molecular tasks. By meticulously orchestrating these settings, we ensure that GIT-Former can harness its full potential in varied application scenarios.

\begin{itemize}
\item \textbf{Pre-training and Hyperparameters}: The encoder settings freeze both the image and graph encoders. For modality selection, 1-3 arbitrary modalities and the target text modality are selected for pre-training. The learning rate ranges from 1e-4 to 5e-5, and the training process is constrained to less than 20 epochs. The batch size is determined by the number of input modalities, ranging from 32 to 128 (the more input modalities, the smaller the batch size).
\item \textbf{Optional Settings}: There are options for replacement in the GIT-Former model, such as replacing the Embedding layer with other models like BERT or MolT5's Encoder. Training strategies can also be tailored to the experiment, including Xmodal-Text Matching (XTM), Xmodal-Text Contrastive Learning (XTC), or both.
\end{itemize}

\begin{itemize}
\item \textbf{Molecular Translation Task}: The task type determines the hyperparameters, including input and output modalities and batch size. The model has No parameters frozen, and the learning rate ranges from 1e-4 to 5e-5. The model is trained for less than 20 epochs, with a batch size between 32 and 64.
\item \textbf{Molecular Property Prediction Task}: The task determines specific hyperparameters, such as batch size. The model's Embedding layer utilizes MolT5-large. The learning rate is again set between 1e-4 and 5e-5, with less than 100 epochs in training. Patience is set to 10, and batch size ranges between 16 and 256.
\end{itemize}

\section{Discussion}
\label{sec:Discussion}

\textbf{Data Quality:} In molecular science, the quality and diversity of data are crucial. Multi-modal data provides a more comprehensive perspective, enhancing model predictions. Furthermore, the datasets require benchmarks, validation subsets, and metrics to gauge their representativeness and challenges.

\textbf{Model Techniques:} Large language models, like the GPT series, have achieved significant success in natural language processing and multi-modal tasks. Applying pre-trained models to specific molecular tasks and refining them can be highly beneficial. Moreover, distillation techniques can shrink models, fitting them for limited-resource settings, and data distillation targets training valuable subsets from large datasets. On the other hand, incorporating molecular science knowledge into model structures can pave a more robust learning pathway, demanding specially designed model layers or preprocessing steps.

\textbf{Scientific Insights:} Beyond vector visualization, visual outputs from models can aid scientists in comprehending decision-making processes. Attention mechanisms and other interpretability tools can reveal the data parts the model relies on for predictions and which structures in the model focus on specific knowledge areas.

\textbf{\added{Practical Implications:}} \added{The improvement of 5\%-10\% in accuracy for property prediction and a significant 20.2\% enhancement in molecule generation validity demonstrate the superior generalizability of our multi-modal approach.
The advancements benefit AI-aided drug discovery (AIDD). A model with higher predictive precision can speed up the process of drug discovery, enhancing both time and cost efficiency.
In drug discovery, where expenses can reach billions, even a 5\% increase in accuracy could mean saving millions, marking a substantial economic impact.
}

\section{Limitations}
\label{sec:Limitations}
\added{
While our model leverages the parameter-efficient training method of prompt learning, the speed of training is still a challenge.
To address this, especially as tasks become complex and the model size grows, we're considering the use of Parameter-Efficient Fine-Tuning (PEFT) to reduce the number of training parameters and enhance performance.
However, for tasks like molecular captioning and generation, we face another hurdle: the absence of appropriate evaluation to measure the model's ability.
This limitation makes it challenging to assess the model's effectiveness in these specific areas.
}

\section{Conclusions}
\label{sec:Conclusions}
% ...
We introduce GIT-Mol, a specialized multi-modal large language model tailored for molecular science, \added{overcoming significant challenges in data collection, particularly in acquiring scarce molecular captions. Additionally, we faced and overcame difficulties in training large models, employing advanced parallel training algorithms.}
Compared to baseline models, it demonstrates relative advantages in tasks such as molecule captioning, generation, image captioning, and property prediction. In addition, a new multi-modal data mixer, GIT-Former, is proposed for fusing multi-modal molecular data. Our model outperforms existing methods and offers an any-to-language modality translation strategy, enhancing flexibility for various applications.

\added{For future directions, we plan to integrate techniques such as Low-Rank Adaptation (LoRA) and Adapters to expedite the training process.
Moreover, we will explore the extent to which prompts enhance model performance.
As for new downstream tasks, such as compound name recognition and chemical reaction prediction, we will enhance the model's flexibility and adaptability.
These improvements are expected to significantly impact fields like molecular science and drug development, potentially leading to more efficient and insightful research.
}
\section*{Data and Software Availability:}
The data and software can be accessed at \url{https://github.com/AI-HPC-Research-Team/GIT-Mol}. It is available for non-commercial use.

\section*{Funding}
This work is supported by grants from the National Natural Science Foundation of China (61902446, 62172456, and 91937302) and the Peng Cheng Cloud-Brain of Peng Cheng Laboratory.

\section*{CRediT authorship contribution statement}
Pengfei Liu: Conceptualization, Methodology, Data curation, Model traning, Writing-original draft.
Yiming Ren: Model traning, Writing-original draft.
Jun Tao: Writing review editing.
Zhixiang Ren: Conceptualization, Formal analysis, Supervision, Funding acquisition, Writing review editing.

\section*{Declaration of Competing Interest}
The authors declare that they have no known competing financial interests or personal relationships that could have appeared to influence the work reported in this paper.

\section*{Acknowledgments} 
The authors appreciate Yue Zhou from Peng Cheng Laboratory for the technical advice. The research is supported by the Peng Cheng Cloud-Brain of Peng Cheng Laboratory.

\printcredits

\bibliographystyle{cas-model2-names}

\bibliography{cas-refs}


\end{document}